\documentclass[11pt]{article}

\usepackage[]{acl}
\usepackage{multirow}
\usepackage{times}
\usepackage{booktabs}
\usepackage{tikz}
\usetikzlibrary{positioning}
\usepackage{graphicx}

\usetikzlibrary{positioning, arrows.meta}
\usepackage{subcaption}

\usepackage{latexsym}
\usepackage{amsmath}
\usepackage{amssymb}
\usepackage{tikz}
\usetikzlibrary{arrows.meta, positioning}
\usepackage{comment}
\usepackage[T1]{fontenc}

\usepackage[utf8]{inputenc}

\usepackage{microtype}

\usepackage{inconsolata}

\usepackage{graphicx}
\graphicspath{{figs/}{}}
\usepackage{listings}
\usepackage{xcolor}

\definecolor{codegray}{rgb}{0.95,0.95,0.95}
\definecolor{codeblue}{rgb}{0.2,0.2,0.7}
\definecolor{codegreen}{rgb}{0,0.5,0}
\definecolor{codered}{rgb}{0.7,0.1,0.1}

\lstdefinestyle{mystyle}{
    backgroundcolor=\color{codegray},
    commentstyle=\color{codegreen},
    keywordstyle=\color{codeblue},
    stringstyle=\color{codered},
    basicstyle=\ttfamily\small,
    breakatwhitespace=false,
    breaklines=true,
    captionpos=b,
    keepspaces=true,
    numbers=none,
    showspaces=false,
    showstringspaces=false,
    showtabs=false,
    tabsize=2,
    frame=single,
    frameround=tttt
}

\title{
DreamerNLplus: Interpretable Modeling of Mental Health Dynamics from Social Media Timelines using Hybrid Rule-Based and RAG Methods
}

\author{ Maryia Zhyrko$^{1\dagger}$, Daisy Monika Lal$^{2\dagger}$, Erik van Mulligen$^{3}$, Lifeng Han$^{1,4}$
 \\ On behalf of the 4D PICTURE consortium \\
  $^1$Leiden Institute of Advanced Computer Science (LIACS), Leiden University, NL\\
  $^2$School of Computing and Communications (SCC),
Lancaster University, UK \\
  $^3$Department of Medical Informatics, Erasmus University Medical Center Rotterdam, NL\\
  $^4$Biomedical Data Sciences, Leiden University Medical Center, NL \\
$^{\dagger}$ co-first, corresponding: \{l.han\} @ lumc.nl  \\
 \\
 \\
}

\begin{document}
\maketitle
\begin{abstract}



%
We present DreamerNLplus, a hybrid framework for modeling mental health dynamics from social media timelines in the CLPsych 2026 shared task. Our system addresses three tasks: psychological state modeling, temporal change detection, and sequence-level summarization.
For Task 1, we combine LLM-based data augmentation, DeBERTa classification, and Random Forest regression for structured state prediction. For Task 2, we use few-shot prompting with a locally deployed Llama 3.1 model to detect Switch and Escalation events using short-term temporal context. 
For Task 3.1, we explore both a deterministic rule-based summarization pipeline and a few-shot LLM-based approach, ranking \textbf{2nd} officially.
Our RAG-based method achieves strong performance in Task 3.2, ranking \textbf{1st} for Improvement and \textbf{3rd} for Deterioration, demonstrating its ability to capture recurrent psychological change patterns across timelines. 
Our analysis reveals key challenges, including the mismatch between classification and regression performance, the difficulty of modeling temporal transitions, and the disagreement between semantic and similarity-based evaluation metrics. 
These findings highlight the complexity of modeling mental health dynamics and motivate future work on unified evaluation frameworks.
We share our code and prompts at \url{https://github.com/4dpicture/CLPsych2026}

\end{abstract}

\section{Introduction and Background}
We describe our system submissions to the CLPsych 2026 Shared Task \cite{ali026overview}: Capturing and Characterizing Mental Health Changes through Social Media Timeline Dynamics, where we have attended all the sub-tasks.
Earlier reviews on NLP for mental health research 
\cite{le2021machine,malgaroli2023natural} provide an overview of traditional NLP and ML methods including rule-based, statistical (TF-IDF, decision trees, SVMs, CRFs, random forests, NNs), pretrained LMs (BERT-like encoder models, sequence-to-sequence BART models, domain-specific MentalBERT), and early generative decoders (e.g., GPT2). 
They also pointed out the \textit{limitations} of existing works such as low reproducibility issue. The resources used include electronic health records (EHRs), Psychological Evaluation reports, social media and interview data.
For this study, we used social media post data from the shared task.
To improve model \textit{interpretability} as well as investigating newer open-source models, we applied a hybrid combination of a rule-based method, a random forest classifier, DeBERTa models, RAG, and open-sourced LLMs.
Recent work related to ours include RAG \cite{kermani-etal-2025-systematic,bogdanova2026flans}, 
prompt engineering and PA-ISP (perspective-aware)
\cite{chan-etal-2025-prompt,romero2025manchester,ren2025malei}, and CLPsych systems/dataset \cite{tseriotou-etal-2025-overview,atzil_slonim_2025_mind,tsakalidis-etal-2022-overview,atzil_slonim_2026}.

The three main tasks and their sub-tasks:\footnote{
Task 1: \url{https://www.codabench.org/competitions/14057/}
Task 2: \url{https://www.codabench.org/competitions/14703/}
Task 3: \url{https://www.codabench.org/competitions/14669/}
}

\begin{itemize}
    \item \textbf{Task 1:} Predict adaptive and maladaptive ABCD element combinations:
    (1.1) Post-level identification of dominant ABCD subelements and self-state composition;
    (1.2) Self-state presence rating.
    
    \item \textbf{Task 2:} Identify moments of change.
    
    \item \textbf{Task 3:} Summary of change:
    (3.1) Summarizing sequences surrounding change events;
    (3.2) Identifying recurrent dynamic signatures of change across timelines.
\end{itemize}









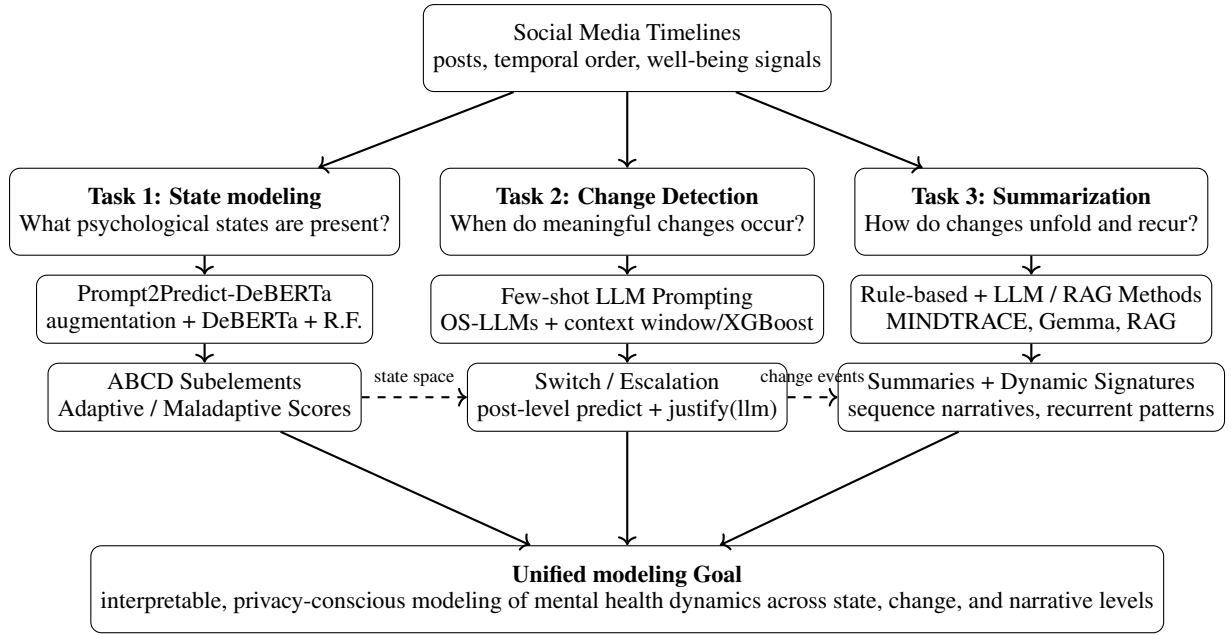
\begin{figure*}[t]
\centering
\begin{tikzpicture}[
    node distance=.25cm,
    every node/.style={font=\footnotesize},
    taskbox/.style={rectangle, draw, rounded corners, align=center,
                    minimum width=2.2cm, minimum height=1.15cm},
    smallbox/.style={rectangle, draw, rounded corners, align=center,
                    minimum width=1.7cm, minimum height=0.9cm},
    arrow/.style={->, thick}
]

\node[taskbox, minimum width=2.5cm] (input) {Social Media Timelines\\
\footnotesize posts, temporal order, well-being signals};

\node[taskbox, below left=1.cm and .3 cm of input] (task1) {
\textbf{Task 1: State modeling}\\
\footnotesize What psychological states are present?
};

\node[smallbox, below=of task1] (t1method) {
Prompt2Predict-DeBERTa\\
\footnotesize  augmentation + DeBERTa + R.F.
};

\node[smallbox, below=of t1method] (t1out) {
ABCD Subelements\\
Adaptive / Maladaptive Scores
};

\node[taskbox, below=1.cm of input] (task2) {
\textbf{Task 2: Change Detection}\\
\footnotesize When do meaningful changes occur?
};

\node[smallbox, below=of task2] (t2method) {
Few-shot LLM Prompting\\
\footnotesize OS-LLMs + context window/XGBoost
};

\node[smallbox, below=of t2method] (t2out) {
Switch / Escalation \\
\footnotesize post-level predict + justify(llm)
};

\node[taskbox, below right=1.cm and .3 cm of input] (task3) {
\textbf{Task 3: Summarization}\\
\footnotesize How do changes unfold and recur?
};

\node[smallbox, below=of task3] (t3method) {
Rule-based + LLM / RAG Methods\\
\footnotesize MINDTRACE, Gemma, RAG
};

\node[smallbox, below=of t3method] (t3out) {
Summaries + Dynamic Signatures\\
\footnotesize sequence narratives, recurrent patterns
};

\draw[arrow] (input) -- (task1);
\draw[arrow] (input) -- (task2);
\draw[arrow] (input) -- (task3);

\draw[arrow] (task1) -- (t1method);
\draw[arrow] (t1method) -- (t1out);

\draw[arrow] (task2) -- (t2method);
\draw[arrow] (t2method) -- (t2out);

\draw[arrow] (task3) -- (t3method);
\draw[arrow] (t3method) -- (t3out);

\draw[arrow, dashed] (t1out.east) -- node[above, font=\scriptsize] {state space} (t2out.west);
\draw[arrow, dashed] (t2out.east) -- node[above, font=\scriptsize] {change events} (t3out.west);

\node[taskbox, below=1.5cm of t2out, minimum width=10.5cm] (summary) {
\textbf{Unified modeling Goal}\\
\footnotesize interpretable, privacy-conscious modeling of mental health dynamics across state, change, and narrative levels
};

\draw[arrow] (t1out) -- (summary);
\draw[arrow] (t2out) -- (summary);
\draw[arrow] (t3out) -- (summary);

\end{tikzpicture}

\caption{Overview of the DreamerNLplus system across the CLPsych 2026 shared tasks. Task 1 models psychological state representations, Task 2 detects temporal moments of change, and Task 3 summarizes and generalizes change dynamics across sequences.}
\label{fig:system_overview}
\end{figure*}

\section{DreamerNLplus Methods}

Figure~\ref{fig:system_overview} summarizes the overall DreamerNLplus framework. Across the three shared tasks, our systems follow a unified modeling perspective: Task 1 identifies psychological state representations, Task 2 detects transitions between states, and Task 3 converts these dynamics into sequence-level summaries and recurrent signatures.











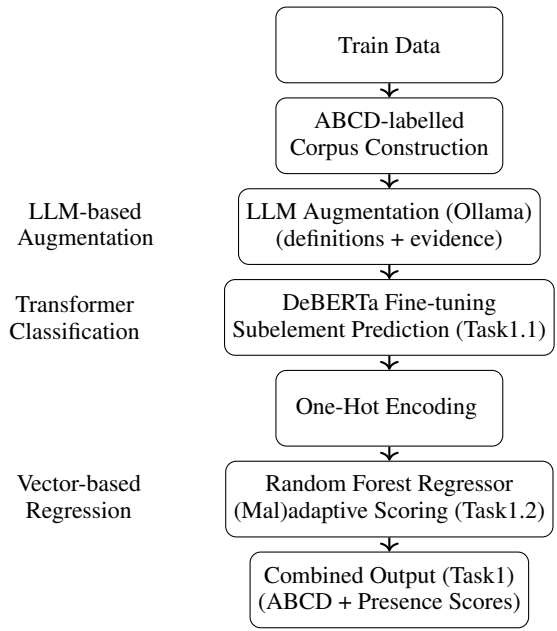
\begin{figure}[t]
\footnotesize
\centering
\begin{tikzpicture}[
    node distance=1.2cm,
    every node/.style={font=\footnotesize},
    box/.style={rectangle, draw, rounded corners, align=center, minimum width=3cm, minimum height=1cm},
    arrow/.style={->, thick}
]

\node[box] (data) {Train Data};
\node[box, below of=data] (corpus) {ABCD-labelled\\Corpus Construction};

\node[box, below of=corpus] (augment) {LLM Augmentation (Ollama)\\
\footnotesize (definitions + evidence)};

\node[box, below of=augment] (deberta) {DeBERTa Fine-tuning\\
Subelement Prediction (Task1.1)};

\node[box, below of=deberta] (vector) {One-Hot Encoding
};

\node[box, below of=vector] (rf) {Random Forest Regressor\\
(Mal)adaptive Scoring (Task1.2)};

\node[box, below of=rf] (output) {Combined Output (Task1)\\
(ABCD + Presence Scores)};

\draw[arrow] (data) -- (corpus);
\draw[arrow] (corpus) -- (augment);
\draw[arrow] (augment) -- (deberta);
\draw[arrow] (deberta) -- (vector);
\draw[arrow] (vector) -- (rf);
\draw[arrow] (rf) -- (output);

\node[left=1cm of augment, align=center] {LLM-based\\Augmentation};
\node[left=1cm of deberta, align=center] {Transformer\\Classification};
\node[left=1cm of rf, align=center] {Vector-based\\Regression};

\end{tikzpicture}

\caption{
Prompt2Predict-DeBERTa pipeline including data preprocessing for Task 1 (Predict adaptive and maladaptive ABCD element combinations). 
}
\label{fig:task1_pipeline}
\end{figure}

\begin{figure*}[t!]
    \centering
    \includegraphics[width=0.9\linewidth]{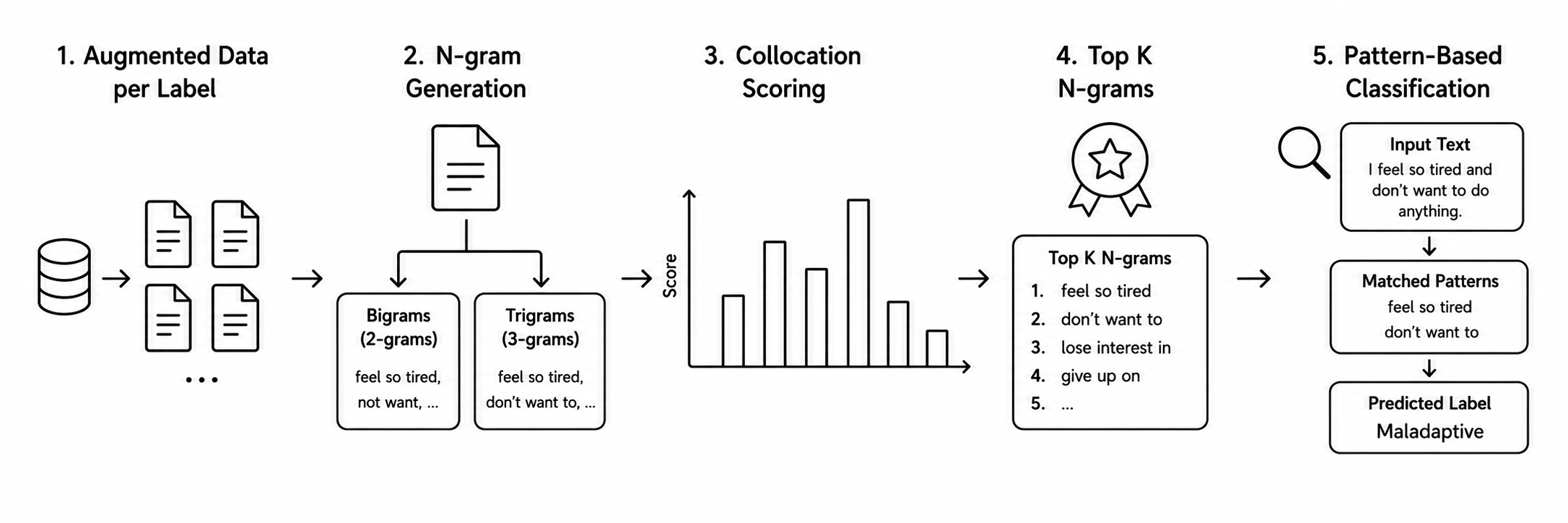}
    \caption{Task 1 rule-based pattern matching approach using n-gram collocations to classify post sentences into ABCD sub-categories.}
    \label{fig:ngram}
\end{figure*}

\textbf{{Tasks 1 and 2 on State Modeling and Change Detection.}}
For \textbf{Task-1}, we propose \textbf{Prompt2Predict-DeBERTa}, a simple multi-stage framework for predicting psychological
sub-elements and presence scores, as in Figure \ref{fig:task1_pipeline}. 
First, we extract evidence for each label from the original training data and augment the evidence using synthetic data for model training purposes.
To do this, we employ Ollama to expand the dataset by {\textit{generating new examples}} through prompts that include label definitions and annotated evidence for every ABCD category.
The augmented data is used for DeBERTa model fine-tuning on the task of subelement prediction (Task1.1). 
The prediction output (ABCD labels) will be encoded as one-hot vector and fed as input to the Random Forest Regressor to further predict the Adaptive / Maladaptive scoring (Task1.2). 
Finally, we combine both outputs to include two elements: ABCD labels and their presence ratings.
We also deployed a rule-based approach for Task1, as in Figure \ref{fig:ngram}.

This pipeline combines LLM-based augmentation (data preprocessing), transformer-based classification, and lightweight vector-based regression to produce structured and interpretable predictions, as well as comparing well-defined rules.

\begin{figure}[htbp]
\centering
\begin{tikzpicture}[
    node distance=1.1cm,
    every node/.style={font=\small},
    box/.style={rectangle, draw, rounded corners, align=center,
                minimum width=3.5cm, minimum height=0.9cm},
    smallbox/.style={rectangle, draw, rounded corners, align=center,
                minimum width=2.7cm, minimum height=0.8cm},
    arrow/.style={->, thick}
]

\node[box] (json) {Timeline JSON};

\node[box, below of=json] (context) {Context Window Formatting\\
\footnotesize current post + up to 5 preceding posts};

\node[box, below of=context] (prompt) {5-shot Prompt Construction\\
\footnotesize Switch + Escalation examples};

\node[box, below of=prompt] (llm) {Local LLM Inference\\
\footnotesize Llama 3.1 via Ollama};

\node[box, below of=llm] (pred) {Post-level Prediction\\
\footnotesize Switch / Escalation binary labels};

\node[box, below of=pred] (justify) {Structured Output\\
\footnotesize labels + justification};

\node[box, below of=justify] (submit) {Submission Formatting};

\draw[arrow] (json) -- (context);
\draw[arrow] (context) -- (prompt);
\draw[arrow] (prompt) -- (llm);
\draw[arrow] (llm) -- (pred);
\draw[arrow] (pred) -- (justify);
\draw[arrow] (justify) -- (submit);

\node[smallbox, right=.2cm of context] (temporal) {Temporal Context\\
\footnotesize local change signal};
\node[smallbox, right=.3cm of llm] (privacy) {Privacy-preserving\\
\footnotesize local deployment};
\node[smallbox, right=.2cm of pred] (labels) {Targets\\
\footnotesize Switch, Escalation};

\draw[arrow, dashed] (temporal) -- (context);
\draw[arrow, dashed] (privacy) -- (llm);
\draw[arrow, dashed] (labels) -- (pred);

\end{tikzpicture}

\caption{Task 2 few-shot prompting pipeline (Identify moments of change). 
}
\label{fig:task2_pipeline}
\end{figure}
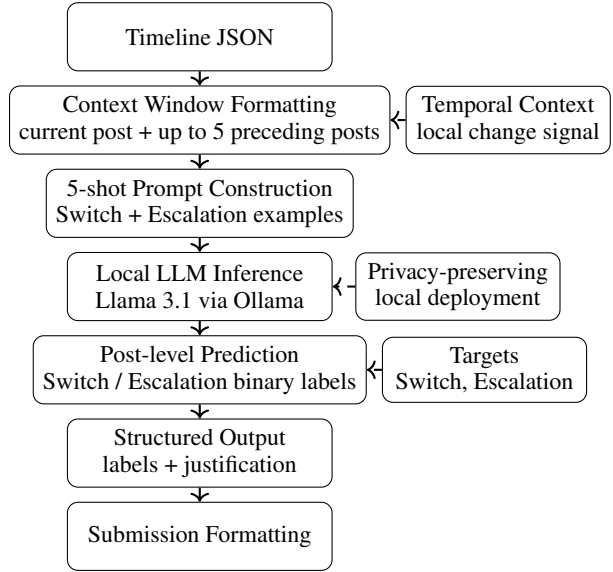

\noindent
\textbf{Task-2: Few-Shot LLM Prompting and XGBoost} 
We first design a framework that supports multiple LLM ++backends (Ollama, HuggingFace) and runs locally for privacy, with Llama 3.1 8B as our submitted model (Figure \ref{fig:task2_pipeline}).
In this framework, we create a local context window by keeping the preceding 5 timeline posts as context to the LLM prompts for predicting targets (Switch and Escalation).
The prompt also defines Switch and Escalation, in addition to the context examples.
Our examples include the combinations of Switch-only, Escalation-only, both, and neither, plus a first-post case where no change is possible (no preceding context). 
The LLM is asked to return its answer in a fixed JSON format with Switch, Escalation, and short justification, and the framework retries the prompt if the response does not match this format. 

As a non-LLM comparison, we build a \textbf{classifier model} using the XGBoost library. 
In the preprocessing of this model, we use TF-IDF and sentence transformer embeddings (all-mpnet-base-v2) to represent the posts.
To get the temporal difference features, we use 1) the embedding difference between the current and previous posts (i.e., current post embedding - previous post embedding $h_t-h_{t-1}$), 2) their element-wise product, and 3) timeline position.
We also used additional feature embeddings, including 14 linguistic features related to sentiment, punctuation, and length (Table \ref{tab:linguistic-features-task2}).
Two binary classifiers then predict Switch and Escalation, with the rare positive examples given more weight during training so the model does not simply default to predicting no change everywhere.\footnote{Positive examples are up-weighted via XGBoost's \texttt{scale\_pos\_weight} parameter, set to $\min\!\left(\frac{n_{\text{neg}}}{n_{\text{pos}}},\ 20\right)$ independently for each classifier.}

\begin{figure*}[t]
\centering
\begin{subfigure}{0.48\textwidth}
\centering
\begin{tikzpicture}[
    node distance=.25cm and .15cm,
    every node/.style={font=\scriptsize},
    box/.style={rectangle, draw, rounded corners, align=center,
                minimum width=2.6cm, minimum height=0.55cm},
    arrow/.style={-{Latex}, thick}
]
\node[box] (input) {Test Data\\MIND-annotated posts};
\node[box, right=of input] (extract) {n-gram / keyword\\ABCD extraction};
\node[box, right=of extract] (state) {Structured State\\Representation};
\node[box, below=of state] (agg) {Presence-based\\Aggregation};
\node[box, left=of agg] (seg) {Temporal Segmentation\\early / middle / late};
\node[box, left=of seg] (change) {Switch \& Escalation\\Detection};
\node[box, below=of change] (dom) {Dominance Scoring\\adaptive vs maladaptive};
\node[box, right=of dom] (theme) {Rule-based\\Theme Inference};
\node[box, right=of theme] (gen) {Template-driven\\Narrative Generation};
\node[box, below=.2cm of gen, minimum width=2.6cm] (summary) {Structured Summary};

\draw[arrow] (input) -- (extract);
\draw[arrow] (extract) -- (state);
\draw[arrow] (state) -- (agg);
\draw[arrow] (agg) -- (seg);
\draw[arrow] (seg) -- (change);
\draw[arrow] (change) -- (dom);
\draw[arrow] (dom) -- (theme);
\draw[arrow] (theme) -- (gen);
\draw[arrow] (gen) -- (summary);
\end{tikzpicture}
\caption{MINDTRACE-SUMMARY: deterministic rule-based pipeline.}
\label{fig:task31_mindtrace}
\end{subfigure}
\hfill
\begin{subfigure}{0.48\textwidth}
\centering
\begin{tikzpicture}[
    node distance=.55cm and .15cm,
    every node/.style={font=\scriptsize},
    box/.style={rectangle, draw, rounded corners, align=center,
                minimum width=2.6cm, minimum height=0.55cm},
    note/.style={rectangle, draw, dashed, rounded corners, align=center,
                minimum width=2.4cm, minimum height=0.55cm, text=gray},
    arrow/.style={-{Latex}, thick}
]
\node[box, minimum width=2.0cm] (posts) {Sequence \\ Posts};
\node[box, right=of posts] (format) {Input Formatting\\text + Switch/Esc.\\+ well-being};
\node[box, right=of format] (examples) {Dynamic Few-shot\\Selection};
\node[box, below=of examples] (prompt) {Prompt Construction\\MIND+ABCD+relational};
\node[box, left=of prompt, minimum width=1.8cm] (llm) {Gemma 2 9B \\ Gemma 4 E4B \\ LLAMA 3.1 8B
\\Inference};
\node[box, left=of llm, minimum width=1.8cm] (out) {Summary \\ Output};

\draw[arrow] (posts) -- (format);
\draw[arrow] (format) -- (examples);
\draw[arrow] (examples) -- (prompt);
\draw[arrow] (prompt) -- (llm);
\draw[arrow] (llm) -- (out);

\node[note, below=.2cm of prompt] (note1) {No Task 1 ABCD\\annotations used};
\node[note, below=.2cm of llm] (note2) {ABCD dynamics\\inferred from text};
\draw[-{Latex}, dashed, gray] (note1) -- (prompt);
\draw[-{Latex}, dashed, gray] (note2) -- (llm);
\end{tikzpicture}
\caption{Few-shot LLM prompting pipeline using OS-LLMs.}
\label{fig:task31_gemma}
\end{subfigure}
\caption{Task 3.1 summary generation methods from DreamerNLplus -- rule-based (left) vs OS-LLMs (right). 
}
\label{fig:task31_methods}
\end{figure*}
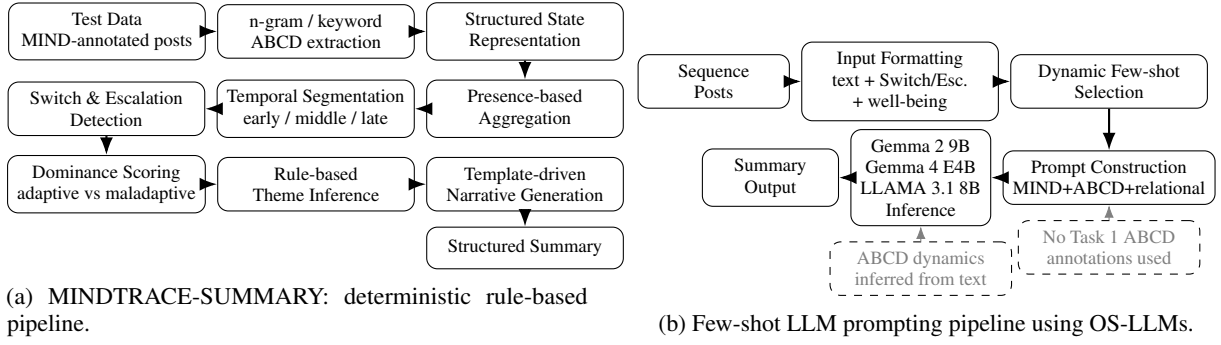

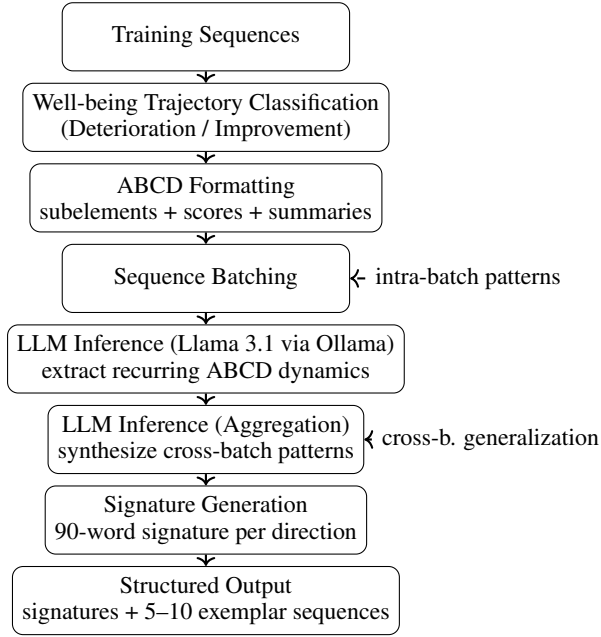
\begin{figure}[t]
\centering
\begin{tikzpicture}[
    node distance=.15cm,
    every node/.style={font=\footnotesize},
    box/.style={rectangle, draw, rounded corners, align=center,
                minimum width=3.8cm, minimum height=0.9cm},
    stage/.style={rectangle, draw, thick, dashed, rounded corners,
                minimum width=4.4cm, minimum height=3.5cm},
    arrow/.style={->, thick}
]

\node[box] (input) {Training Sequences};

\node[box, below=of input] (traj) {Well-being Trajectory Classification\\
\footnotesize (Deterioration / Improvement)};

\node[box, below=of traj] (format) {ABCD Formatting\\
\footnotesize subelements + scores + summaries};


\node[box, below=0.15cm of format] (batch) {Sequence Batching};

\node[box, below=of batch] (llm1) {LLM Inference (Llama 3.1 via Ollama)\\
\footnotesize extract recurring ABCD dynamics};


\node[box, below=of llm1] (llm2) {LLM Inference (Aggregation)\\
\footnotesize synthesize cross-batch patterns};

\node[box, below=of llm2] (sig) {Signature Generation\\
\footnotesize 90-word signature per direction};

\node[box, below=of sig] (output) {Structured Output\\
\footnotesize signatures + 5--10 exemplar sequences};

\draw[arrow] (input) -- (traj);
\draw[arrow] (traj) -- (format);
\draw[arrow] (format) -- (batch);
\draw[arrow] (batch) -- (llm1);
\draw[arrow] (llm1) -- (llm2);
\draw[arrow] (llm2) -- (sig);
\draw[arrow] (sig) -- (output);

\node[right=.2cm of batch] (side1) {\footnotesize intra-batch patterns};
\draw[arrow, dashed] (side1) -- (batch);

\node[right=.1cm of llm2] (side2) {\footnotesize cross-b. generalization};
\draw[arrow, dashed] (side2) -- (llm2);
\end{tikzpicture}
\caption{Overview of the RAG-LLM Signature Mining framework for Task 3.2. 
}
\label{fig:task32_pipeline}
\end{figure}

\noindent
\textbf{{Task 3 on Change Summarization and Pattern Mining.}}
For \textbf{Task 3.1} ``summarizing sequences surrounding change events'', we explore two distinct summary generation strategies (Figure~\ref{fig:task31_methods}).

\noindent
\textbf{MINDTRACE-SUMMARY}
is a fully deterministic, multi-stage framework for generating
sequence-level psychological signatures from MIND-annotated posts.
The model generates summaries by following a fixed structure and converting ABCD annotations into
natural language. A full sample template is provided in Appendix~\ref{app: Task3.1 Summary Generation Template}. It first determines whether adaptive or maladaptive states dominate, then builds a
narrative with five parts: central theme, initial state, interaction dynamics, transition (switch or escalation), and outcome. ABCD labels are rewritten into fluent psychological descriptions, and the summary emphasises how states reinforce or shift over time. The method prioritizes consistency and structure over free-form generation, ensuring summaries clearly reflect change dynamics across the sequence.

\noindent
For \textbf{Few-Shot LLM Prompting},
the prompt specifies the MIND framework, ABCD abbreviation conventions, relational dynamics vocabulary, and required summary structure covering pre-change phase, within/between-state dynamics, and explicit change event identification. Task 1 ABCD annotations are not used; the model infers ABCD dynamics from post text guided by the prompt.

For \textbf{Task 3.2} ``identifying recurrent dynamic signatures of change across timelines'',
we propose \textbf{RAG-LLM Signature Mining}, a two-stage LLM-based framework for identifying recurrent dynamic signatures of psychological change.
As illustrated in Figure~\ref{fig:task32_pipeline}, our framework separates intra-batch pattern extraction from cross-batch signature synthesis, enabling the identification of recurrent psychological dynamics across timelines.
Sequences are batched and fed to an open-source LLM (Llama 3.1via Ollama) with per-post ABCD subelements, well-being scores, and gold summaries. Stage 1 extracts recurring ABCD dynamics per batch; Stage 2 synthesizes these into one 90-word signature per direction (deterioration/improvement), with 5–10 exemplar sequences as evidence.


\section{Results and Analysis}

\begin{figure}[t!]
\centering 
\includegraphics[width=.49\textwidth]{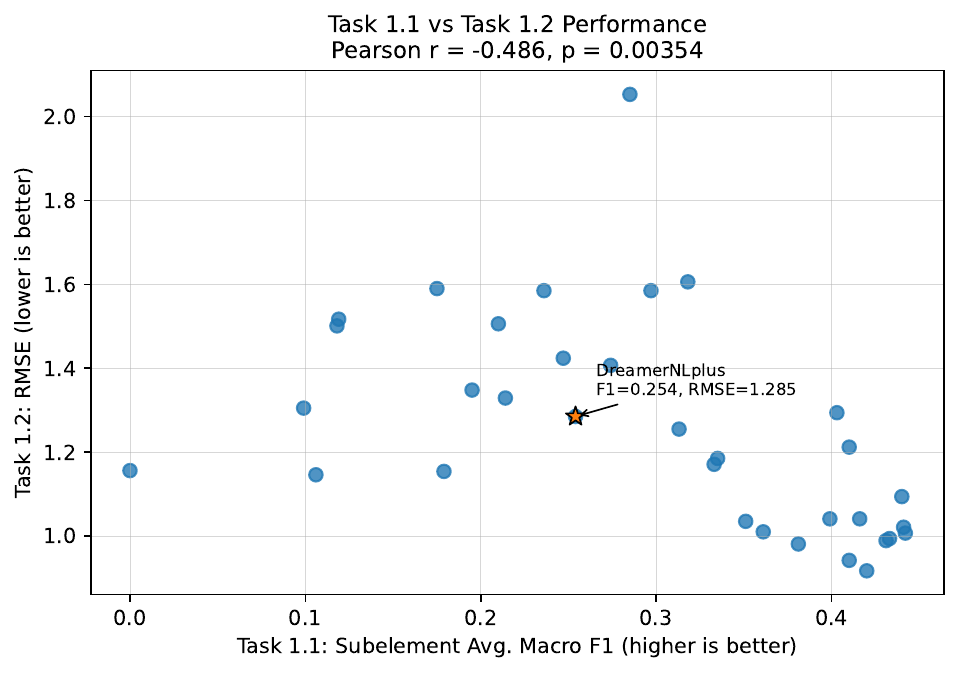}
  \caption{Task 1.1 vs 1.2 Relation. Each point = one team.
X-axis = Task 1.1 (F1) → higher is better.
Y-axis = Task 1.2 (RMSE) → lower is better.
  } 
  \label{fig:task-1.1-vs-1.2-relation}
\end{figure}

\paragraph{Task 1: State modeling}
For Task 1, our system ranks 22nd in Task 1.1 (subelement classification) and 20th in Task 1.2 (presence estimation), as shown in Table \ref{tab:task1_combined_ranking}. 
To further understand this discrepancy, we analyze the relationship between classification performance and regression performance in Figure~\ref{fig:task-1.1-vs-1.2-relation}. The results show a moderate negative correlation ($r = -0.486$, $p = 0.00354$), indicating that stronger subelement classification performance does not necessarily translate into better presence estimation.
This observation reflects the fundamental difference between the two subtasks: Task 1.1 requires fine-grained categorical prediction of psychological subelements, while Task 1.2 evaluates continuous intensity estimation. Our system, which combines DeBERTa-based classification with Random Forest regression, appears more robust in the regression setting, suggesting that downstream aggregation mitigates upstream classification errors.


\paragraph{Task 2: Change Detection}
Our submitted LLM system reaches a combined F1 of 0.442, ranking 11th overall (Figure~\ref{fig:task-2-scores-rank}), while the XGBoost variant scores 0.327. The two approaches show opposite error patterns: the LLM has high recall but low precision (Switch 0.762/0.302, Escalation 0.917/0.393) meaning it picks up most change cues but tends to over-predict them, while XGBoost has high precision but low recall on rare positive classes(Switch precision 0.455, recall 0.238), reflecting how hard it is to learn these labels from only 30 gold timelines.
This shows that accurately capturing subtle transitions remains challenging across paradigms, and that errors on weak or ambiguous change signals can propagate across the sequence.
Despite not relying on task-specific fine-tuning, the LLM method stays competitive performance while providing interpretable predictions with textual justifications. This highlights the effectiveness of few-shot prompting for modeling temporal dynamics in low-resource settings.

\paragraph{{Task 3: Summarization and Pattern Mining}}
For Task 3.1, in our processing, we kept other teams best average-ranked submission and our best-performing submission (ID 693964 prompting-w-ablation) ranks 4th overall based on the average ranking across multiple metrics (Figure~\ref{fig:task-3.1test-all-teams}). However, a notable observation is the strong disagreement between evaluation metrics.


Interestingly, shown in Figure~\ref{fig:task-3.1test-all-teams}, the rule-based submission (694142) achieves the best CT and strong CS performance, yet ranks worst overall due to poor ROUGE and BERT scores (a detailed discussion of the evaluation is provided in Appendix~\ref{app:evaluation rule-based summarization}.), highlighting a clear disagreement between semantic coherence metrics and similarity-based metrics: 
CS and CT emphasize semantic coherence and psychological consistency, whereas ROUGE and BERTScore prioritize surface-level similarity to reference summaries. As a result, optimizing for one set of metrics may degrade performance on the other.
%
%
However, in the official Task 3.1 ranking, DreamerNLplus ranks \textbf{2nd} overall with submission 693964, based on the average of metric-specific ranks across CS, CT, ROUGE-L Recall, and BERTScore Recall.

For Task 3.2, as shown in Table~\ref{tab:task32_results}, our RAG-based approach ranks \textbf{1st} on Improvement and \textbf{3rd} on Deterioration. In addition, our system achieves the highest score on Specificity and the second-highest on Recurrence for Improvement, and the second-highest Specificity for Deterioration.
These results suggest that the proposed RAG-based framework is particularly effective at identifying precise and recurring dynamic patterns across timelines. By combining batch-level pattern extraction with cross-batch synthesis, the approach is able to capture higher-level psychological change signatures that generalize across individuals. However, the performance gap between Improvement and Deterioration also indicates that modeling deterioration patterns remains more challenging, potentially due to greater variability or ambiguity in negative self-state dynamics.
This highlights the importance of modeling both intra-sequence consistency and inter-sequence generalization when identifying recurrent psychological dynamics.

\paragraph{Cross-Task Analysis}

Across all tasks, a consistent pattern emerges: different evaluation metrics capture distinct and sometimes conflicting aspects of performance. Task 1 reveals a trade-off between classification accuracy and regression stability, Task 2 highlights the difficulty of balancing local and temporal consistency, and Task 3 exposes a mismatch between semantic coherence and lexical similarity metrics.
These findings suggest that modeling mental health dynamics requires balancing multiple objectives, including structured prediction, temporal reasoning, and semantic generation. Our hybrid approach demonstrates that combining interpretable representations with flexible LLM-based methods can provide robust performance across tasks, while also revealing important limitations in current evaluation frameworks.

\section{Conclusions and Future Work}




 



We presented \textbf{DreamerNLplus}, a hybrid framework for modeling mental health dynamics from social media timelines across three tasks: psychological state modeling, temporal change detection, and sequence-level summarization. By combining structured representations, few-shot prompting, and rule-based generation, our approach provides both interpretability and flexibility across different modeling paradigms.
Future work should aim to design unified evaluation frameworks that reconcile semantic fidelity, temporal consistency, and textual similarity, enabling  clinically meaningful assessment of mental health modeling systems.




\section*{Limitations}

For data augmentation in Task 1, in this work, we extracted the evidence from original training data, and only asked LLMs to generate more evidence, subsequently, DeBERTa model is trained on the augmented evidence data to predict sub-element classes. This is similar to the dense prescription generation work in \cite{belkadi2025lt3} for NER and engineering purposes only, without generating full clinical letters. In an ideal situation, we will further explore the generation of similar post-level, not only evidence. 

Task 3 template-based summarization achieved high CS and low CT scores through predefined linguistic rules and structured feature-to-text mappings, enabling stable and interpretable summaries. However, the approach is task-specific and less flexible than human summarization, limiting its ability to capture nuanced contextual and emotional variations in narratives. Despite this, it is well-suited for formal or high-stakes settings where standardized and reproducible documentation is prioritized over linguistic diversity. Additionally, the framework relies on intermediate computations such as switch and escalation scores derived from upstream predictions, meaning errors in earlier stages may propagate into the final summaries.


\section*{Ethics}
The shared task data we used in this paper is anonymized and annotated by CLPych2026 organizers. We only used secure methods and models to process the data, such as rule-based, locally hosted open-source LLMs, locally trained encoders, without releasing the data to any third parties with our best practice for privacy protection. 

\section*{Acknowledgments}


We thank David Lindevelt for the help on this project, including the automated prompt pipeline and framework adapted from another project \cite{han2025dutch}. 
We thank Prof. Suzan Verberne for editing the camera-ready version of this paper.
We thank the reviewers for their valuable comments on our work.
We thank the organizers preparing this shared task, and we are grateful for their communication during our registration and submissions, especially Talia Tseriotou
and Iqra Ali from Queen Mary University of London.
The 4D PICTURE consortium is funded by the European Union under Horizon Europe Work Programme 101057332. Views and opinions expressed are however those of the author(s) only and do not necessarily reflect those of the European Union or the European Health and Digital Executive Agency (HaDEA). Neither the European Union nor the granting authority can be held responsible for them.
The UK team are funded under the Innovate UK Horizon Europe Guarantee Programme, UKRI Reference Number: 10041120.

\bibliography{custom}

\clearpage

\appendix

 \clearpage

\section{CLPsych 2026 Shared Tasks}

CLPsych2026 is affiliated with The Workshop on Computational Linguistics and Clinical Psychology,
a workshop series founded in 2014.
CLPsych 2026 will be held at ACL in San Diego, July 4th, 2026.









\subsection{Task Evaluations}
\label{subsec:evaluation_metrics}
\paragraph{Unified Evaluation Framework}
The CLPsych 2026 shared tasks evaluate complementary aspects of modeling mental health dynamics from social media timelines, spanning classification, regression, temporal change detection, and summarization. Across Tasks 1--3, the evaluation framework progressively moves from local, structured predictions to global, sequence-level reasoning and natural language generation.

\textbf{Task 1 (State modeling).}
Task 1 evaluates the ability to model psychological self-states at the post level. Task 1.1 focuses on discrete classification of ABCD elements and subelements, using macro-averaged F1 scores across adaptive and maladaptive categories. Task 1.2 evaluates continuous presence estimation on a 1--5 scale, using regression metrics such as RMSE, with ranking based on the mean RMSE across valences. Together, these subtasks capture the challenge of jointly modeling structured categorical representations and continuous mental state intensity.

\textbf{Task 2 (Change Detection).}
Task 2 evaluates the detection of temporal change signals, specifically Switch (sudden change) and Escalation (gradual change). Performance is measured using F1 scores at both post-level and timeline-level, with final ranking based on the average of these two perspectives. This design rewards systems that can detect both local transitions and maintain consistency across sequences.

\textbf{Task 3 (Change Summarization and Pattern Mining).}
Task 3 evaluates sequence-level understanding and generation. Task 3.1 assesses the quality of generated summaries using a multi-metric framework including semantic consistency (CS), contradiction (CT), and similarity-based metrics (ROUGE-L and BERTScore), with final ranking based on averaged metric ranks. Task 3.2 focuses on identifying recurrent dynamic patterns across timelines, emphasizing generalization and abstraction of psychological change signatures.

\textbf{Cross-Task Perspective.}
Taken together, the evaluation framework reflects a hierarchy of modeling challenges: Task 1 focuses on \textit{what} psychological states are present, Task 2 focuses on \textit{when} meaningful changes occur, and Task 3 focuses on \textit{how} these changes can be summarized and generalized. Notably, the metrics across tasks capture partially complementary qualities, including classification accuracy, regression stability, temporal consistency, and semantic coherence, highlighting the inherent trade-offs in modeling mental health dynamics.

This unified evaluation highlights the difficulty of aligning discrete classification, continuous estimation, temporal reasoning, and semantic generation within a single modeling framework.

\section{Methods and Experimental Development  - Task1 (\textit{More Details})}

We adopt a multi-paradigm modeling strategy combining rule-based, transformer-based, and LLM-assisted components for the overall shared task, to establish the relevance of the adopted approach for a given task.

\begin{figure}
    \centering
    \includegraphics[width=0.6\linewidth]{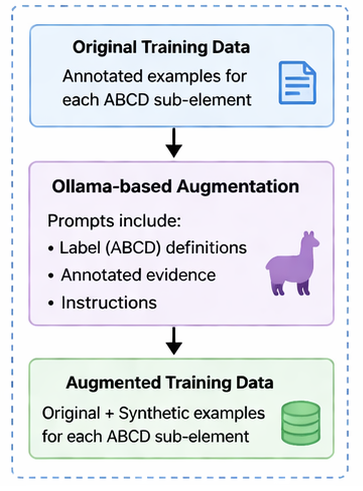}
    \caption{Targeted data augmentation strategy using Ollama.}
    \label{fig:data_augmentaion}
\end{figure}

\subsection{Task 1.1: ABCD Element \& Subelement Classification}
Task 1.1 focuses on identifying fine-grained ABCD self-state elements and their corresponding subelements within each post. However, the task presents two key inherent challenges due to the fine-grained structure of the ABCD schema. First, the large number of subelements per category leads to a highly imbalanced and sparse label space, where certain subelements are significantly under-represented. Second, there exists substantial semantic overlap between closely related subelements, making boundary definition non-trivial even for pretrained language models. 

\begin{figure*}
    \centering
    \includegraphics[width=\linewidth]{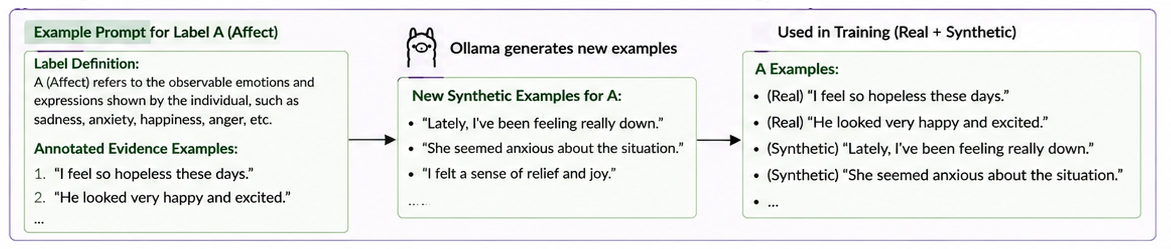}
    \caption{Data Augmentation Examples (paraphrased to preserve privacy in accordance with shared task guidelines).}
    \label{fig:ollamaprompt}
\end{figure*}

\paragraph{Data Augmentation:}
To address data sparsity, we employ targeted data augmentation using Ollamas (see Fig. \ref{fig:data_augmentaion}). Augmentation is guided by structured ABCD element definitions and subelement descriptions, enabling controlled expansion of training data while preserving label semantics (see Fig. \ref{fig:ollamaprompt}). However, augmentation introduces additional challenges. While LLM-based generation improves data volume and diversity, it may also introduce distributional shifts, as synthetic samples often lack the contextual richness and temporal grounding of real social media posts. This is particularly critical in longitudinal mental health modeling, where self-state interpretation depends on subtle linguistic, emotional, and contextual cues.

\paragraph{Transformer-based Approach (DeBERTa):}
Initially, we considered a transformer-based sequence classifier, DeBERTa, for direct multi-label prediction of subelements. The model is fine-tuned to jointly learn element presence and subelement classification in a supervised setting, leveraging contextual embeddings to capture nuanced linguistic signals. However, due to extreme label granularity, semantic overlap between subelements (e.g., self-care vs. self-improvement, anxiety vs. despair), and limited training examples for several classes, purely supervised learning was found to be insufficient for robust generalisation.

\paragraph{Rule-based Approach (Augmented n-gram Modeling):}
Consequently, we introduce a rule-based pipeline that leverages label-conditioned augmented data for n-gram extraction and pattern-based classification (see Fig. \ref{fig:ngram}). Augmented samples generated via Ollama are used to construct label-specific lexical signatures in the form of n-grams, which serve as interpretable indicators of subelement presence. Specifically, we compute top-k bigrams and trigrams from cleaned, stopword-removed text and rank them using likelihood ratio scores to identify statistically salient phrases. This approach is particularly effective in low-resource settings, where explicit lexical cues correlate strongly with specific psychological states.


\subsection{Task 1.2: Presence Rating:}
Task 1.2 focuses on estimating the overall presence rating (1–5) of adaptive and maladaptive states for each post, reflecting the psychological centrality of each state within the narrative. To address this task, we explore both regression-based and LLM-based approaches. We observed a clear relationship between the presence of specific subelements and the final presence ratings, but also noted that simple frequency-based counting of elements did not reliably correspond to the assigned scores. In particular, ratings were influenced not just by the number of subelements but by their specific types within the same ABCD category. 

\paragraph{Regression-based Approach:}
Given the relationship observed between ABCD subelements and presence ratings, we adopt a regression framework using RandomForestRegressor. We construct structured binary feature vectors that serve as inputs to two separate regression models, with adaptive and maladaptive states modelled independently. This approach directly depends on the outputs of Task 1.1 during inference, where detected subelements are aggregated to form the input feature representation. Finally, the continuous outputs are rounded to discrete levels (1–5), aligning with the original annotation scheme.


\subsection{Task 2 Features}

\begin{table*}[ht]
\centering
\footnotesize
\caption{Hand-crafted linguistic features used in Task~2 (Switch/Escalation detection).
         Features are extracted per post and concatenated with the temporal embedding differences.}
\label{tab:linguistic-features-task2}
\begin{tabular}{llp{5.8cm}}
\toprule
\textbf{\#} & \textbf{Feature} & \textbf{Description} \\
\midrule
\multicolumn{3}{l}{\textit{Lexical}} \\
\midrule
1  & \texttt{log\_len}            & $\log(1 + \text{word count})$; compresses post length onto a continuous scale \\
2  & \texttt{n\_sentences}        & Number of sentences, split on \texttt{[.!?]+} \\
3  & \texttt{avg\_word\_len}      & Mean character length across words \\
4  & \texttt{frac\_upper}         & Fraction of characters that are uppercase \\
5  & \texttt{frac\_punct}         & Fraction of characters that are punctuation (\texttt{!?.,;:}) \\
\midrule
\multicolumn{3}{l}{\textit{Punctuation / prosodic signals}} \\
\midrule
6  & \texttt{n\_exclaim}          & Raw count of exclamation marks (\texttt{!}) \\
7  & \texttt{n\_question}         & Raw count of question marks (\texttt{?}) \\
8  & \texttt{n\_ellipsis}         & Raw count of ellipses (\texttt{...}) \\
9  & \texttt{emo\_punct}          & $\min(\texttt{n\_exclaim} + \texttt{n\_question},\;10)$; capped emotional punctuation \\
\midrule
\multicolumn{3}{l}{\textit{Sentiment lexicon}} \\
\midrule
10 & \texttt{frac\_neg}           & Fraction of words matching a negative lexicon (\emph{hate, depressed, pain, alone}, \ldots) \\
11 & \texttt{frac\_pos}           & Fraction of words matching a positive lexicon (\emph{happy, hope, proud, grateful}, \ldots) \\
12 & \texttt{sentiment\_balance}  & $\texttt{frac\_neg} - \texttt{frac\_pos}$; net sentiment polarity \\
\midrule
\multicolumn{3}{l}{\textit{Structural}} \\
\midrule
13 & \texttt{words\_per\_sent}    & $\text{word count} / \text{sentence count}$; average sentence length \\
14 & \texttt{has\_removed}        & Binary flag for \texttt{[removed]} or \texttt{[deleted]} posts \\
\bottomrule
\end{tabular}
\end{table*}

\section{Cross-task Analysis}
From our methods:
Unlike Task 1, which learns explicit ABCD representations, Task 2 relies on contextual prompting to identify dynamic changes directly from timelines.
Together, these approaches reflect a common emphasis on interpretability, temporal sensitivity, and privacy-preserving deployment. Task 1 models what psychological states are present, while Task 2 models when meaningful changes occur.

\subsection{Stratified Sampling and K-fold}

We have tried both Stratified Sampling and K-fold training data split for model development purposes. In the end, we adopted the K-fold approach for our data processing.





 \section{Prompts and Rule-based Templates}
We share our codes and full prompts used for the shared tasks at our Github page \url{https://github.com/4dpicture/CLPsych2026}.

\subsection{Task2 Prompts}

\subsection{Task3.1 Prompts}
\subsection{Task3.1 Summary Generation Template}
\label{app: Task3.1 Summary Generation Template}
We generate the final narrative using a rule-based template conditioned on extracted discourse features. Let $M$ denote the total maladaptive score and $A$ the total adaptive score. Let $\Delta$ denote structural change type and $D$ denote trajectory direction.

\paragraph{Feature Sets}
Adaptive ($\mathcal{F}_a$) and Maladaptive ($\mathcal{F}_m$) features labels.

\paragraph{Initial Phase}
If $M \geq A$, maladaptive processes are dominant:
\begin{quote}
Initially, maladaptive self-state processes are more dominant, characterized by elements such as $\mathcal{F}_m$, while adaptive processes remain less prominent.
\end{quote}
Otherwise, adaptive processes are dominant:
\begin{quote}
Initially, adaptive self-state processes are more dominant, characterized by elements such as $\mathcal{F}_a$, buffering against maladaptive tendencies.
\end{quote}

\paragraph{Temporal Dynamics}
If $M > A$, maladaptive dynamics intensify over time:
\begin{quote}
Maladaptive dynamics intensify over time through reinforcing cycles of negative affect, self-critical cognition, and behavioral withdrawal, suppressing adaptive functioning.
\end{quote}
Otherwise, adaptive processes strengthen over time:
\begin{quote}
Adaptive processes strengthen over time through increasing self-compassion, relational engagement, and constructive coping that counter maladaptive tendencies.
\end{quote}

\paragraph{Structural Transition}
If $\Delta = \text{switch}$:
\begin{quote}
A transition point emerges within the sequence, reflecting a shift in the balance between adaptive and maladaptive self-states.
\end{quote}
If $\Delta = \text{escalation}$:
\begin{quote}
An escalation unfolds across the sequence, reflecting progressive intensification of emotional, cognitive, and behavioural processes over time.
\end{quote}

\paragraph{Trajectory Direction}
If $D = \text{deterioration}$:
\begin{quote}
In the later phase, maladaptive self-state dynamics dominate, reinforcing sustained distress and hopelessness.
\end{quote}
If $D = \text{improvement}$:
\begin{quote}
In the later phase, adaptive self-state dynamics become dominant, supporting resilience and psychological recovery.
\end{quote}
If $D = \text{fluctuation}$:
\begin{quote}
In the later phase, adaptive and maladaptive self-states remain in tension, reflecting ongoing fluctuation between distress and coping.
\end{quote}

\paragraph{Global Template (Always Included)}
Across all sequences, the following integrative statements are included:
\begin{quote}
The central psychological theme across the sequence reflects an evolving interaction between maladaptive distress and adaptive coping processes expressed through affect, cognition, behavior, and desire.
\end{quote}
\begin{quote}
As the sequence progresses, adaptive and maladaptive self-states increasingly interact, creating periods of internal conflict, reflection, and shifting psychological balance.
\end{quote}
\begin{quote}
Across the sequence, adaptive and maladaptive self-states alternate in dominance and suppression, shaping the overall trajectory of psychological change.
\end{quote}

\subsection{Task3.2 Prompts}







\section{Discussion}
In this section, we discuss the performance characteristics, advantages, and limitations of the proposed methods.
\subsection{Consistency vs. Lexical Similarity in Template-Based Summarization}
\label{app:evaluation rule-based summarization}
The template-based summarization method achieved the highest CS and lowest CT scores, highlighting its strength in producing stable and logically coherent outputs. This behavior is expected, as the generation process is strictly governed by predefined linguistic rules and structured feature-to-text mappings, which reduce variability and minimize the risk of semantic drift or internally inconsistent statements. Such control is particularly important in sensitive domains such as palliative care narrative analysis, where reliability and interpretability of generated summaries are critical.

However, this same constrained generation process leads to lower ROUGE and BERTScore performance compared to more flexible neural baselines. Both metrics reward lexical overlap and semantic similarity to reference summaries, which are typically written in a more natural and varied style. Template-based outputs, while semantically faithful and structurally consistent, do not exhibit the paraphrastic richness or lexical alignment required to maximize these scores. As a result, there is an inherent trade-off between consistency-oriented generation and similarity-based evaluation metrics.

This trade-off suggests that template-based summarization is particularly well-suited for applications where factual stability, interpretability, and controllability are prioritized over surface-level similarity to reference texts. In the context of psychological trajectory modeling from patient and caregiver narratives, such methods are advantageous for producing reproducible summaries of adaptive and maladaptive self-state dynamics. Conversely, neural summarization models may be preferred in settings where linguistic expressiveness and alignment with human-written references are more important than strict structural consistency.

\section{Details on rankings and evaluation scores}
We list our official ranking scores with other teams in this section.

\begin{figure}[t!]
\centering 
\includegraphics[width=.49\textwidth]{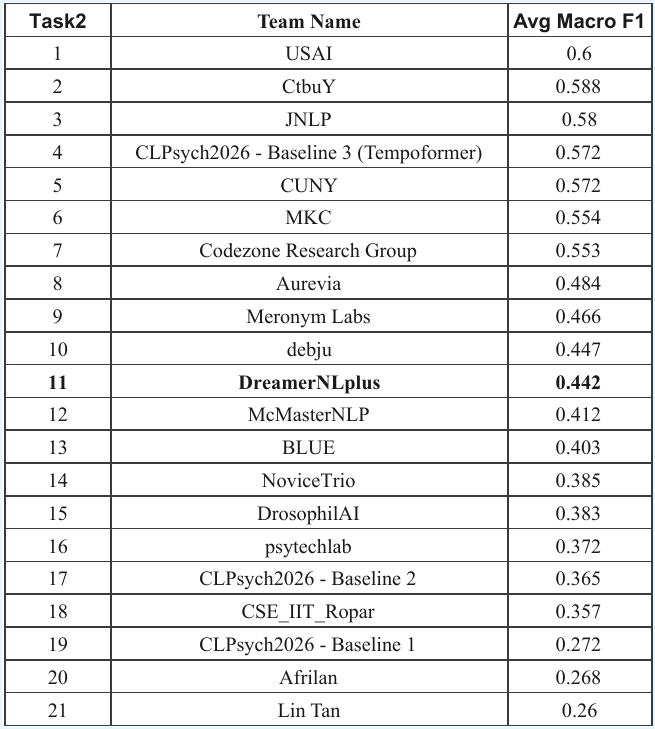}
  \caption{Task 2 Ranking
  } 
  \label{fig:task-2-scores-rank}
\end{figure}


\begin{table*}[t]
\centering
\tiny 
\caption{Task 1.1 and Task 1.2 rankings. Task 1.1 is ranked by Subelement Average Macro F1, while Task 1.2 is ranked by RMSE, where lower is better.}
\label{tab:task1_combined_ranking}
\resizebox{\textwidth}{!}{
\begin{tabular}{lcccc}
\toprule
\textbf{Team} & \textbf{Task 1.1 Rank} & \textbf{Task 1.2 Rank} & \textbf{Task 1.1 Macro F1} & \textbf{Task 1.2 RMSE} \\
\midrule
CUNY & 1 & 6 & 0.442 & 1.007 \\
StateOfMIND & 2 & 8 & 0.441 & 1.021 \\
StateOfMIND & 3 & 12 & 0.440 & 1.094 \\
StateOfMIND & 4 & 5 & 0.433 & 0.994 \\
CUNY & 5 & 4 & 0.431 & 0.989 \\
Meronym Labs & 6 & 1 & 0.420 & 0.917 \\
CUNY & 7 & 10 & 0.416 & 1.041 \\
Meronym Labs & 8 & 18 & 0.410 & 1.212 \\
USAI & 9 & 2 & 0.410 & 0.942 \\
USAI & 10 & 21 & 0.403 & 1.294 \\
Meronym Labs & 11 & 11 & 0.399 & 1.041 \\
Aurevia & 12 & 3 & 0.381 & 0.981 \\
MKC & 13 & 7 & 0.361 & 1.010 \\
McMasterNLP & 14 & 9 & 0.351 & 1.035 \\
ull & 15 & 17 & 0.335 & 1.185 \\
ull & 16 & 16 & 0.333 & 1.171 \\
BLUE & 17 & 33 & 0.318 & 1.606 \\
NoviceTrio & 18 & 19 & 0.313 & 1.255 \\
Afrilan & 19 & 31 & 0.297 & 1.585 \\
Afrilan & 20 & 34 & 0.285 & 2.053 \\
psytechlab & 21 & 25 & 0.274 & 1.407 \\
\textbf{DreamerNLplus} & \textbf{22} & \textbf{20} & \textbf{0.254} & \textbf{1.285} \\
CLPsych2026 - Baseline & 23 & 26 & 0.247 & 1.424 \\
CtbuY & 24 & 30 & 0.236 & 1.585 \\
Afrilan & 25 & 23 & 0.214 & 1.329 \\
CtbuY & 26 & 28 & 0.210 & 1.506 \\
debju & 27 & 24 & 0.195 & 1.348 \\
DrosophilAI & 28 & 14 & 0.179 & 1.154 \\
CSE\_IIT\_Ropar & 29 & 32 & 0.175 & 1.590 \\
ull & 30 & 29 & 0.119 & 1.517 \\
CtbuY & 31 & 27 & 0.118 & 1.501 \\
CSE\_IIT\_Ropar & 32 & 13 & 0.106 & 1.146 \\
DrosophilAI & 33 & 22 & 0.099 & 1.305 \\
debjy & 34 & 15 & 0.000 & 1.156 \\
\bottomrule
\end{tabular}
}
\end{table*}

\begin{table*}[t]
\centering
\footnotesize
\caption{\textbf{The Official rank} of Task 3.1 based on selected submissions of all teams. CS, ROUGE-L Recall, and BERTScore Recall are higher-is-better; CT is lower-is-better. Final rank is based on the average of metric-specific ranks. We rank the \textbf{2nd best} overall.}
\label{tab:task31_official_rank}
\resizebox{\textwidth}{!}{
\begin{tabular}{clcccccccccc}
\toprule
\textbf{Rank} & \textbf{Team} & \textbf{Sub. ID} 
& \textbf{CS} & \textbf{CS Rank} 
& \textbf{CT} & \textbf{CT Rank} 
& \textbf{ROUGE-L} & \textbf{R-L Rank} 
& \textbf{BERT} & \textbf{BERT Rank} 
& \textbf{Avg. Rank} \\
\midrule
1  & MERONYM\_LABS     & 694229 & 0.801 & 3  & 0.659 & 3  & 0.266 & 6  & 0.345 & 4  & 4.00 \\
\textbf{2}  & \textbf{DreamerNLplus} & \textbf{693964} & \textbf{0.735} & \textbf{7}  & \textbf{0.767} & \textbf{7}  & \textbf{0.285} & \textbf{4}  & \textbf{0.345} & \textbf{3}  & \textbf{5.25} \\
3  & CUNY             & 694216 & 0.789 & 5  & 0.714 & 5  & 0.292 & 3  & 0.295 & 9  & 5.50 \\
4  & NoviceTrio       & 693913 & 0.705 & 8  & 0.771 & 8  & 0.318 & 2  & 0.341 & 5  & 5.75 \\
5  & USAI             & 693912 & 0.681 & 10 & 0.849 & 13 & 0.333 & 1  & 0.365 & 1  & 6.25 \\
5  & Aurevia          & 693454 & 0.866 & 1  & 0.625 & 2  & 0.185 & 11 & 0.226 & 11 & 6.25 \\
7  & MKC              & 687777 & 0.654 & 11 & 0.834 & 10 & 0.284 & 5  & 0.359 & 2  & 7.00 \\
8  & psytechlab       & 689973 & 0.857 & 2  & 0.571 & 1  & 0.078 & 13 & 0.147 & 13 & 7.25 \\
9  & JNLP             & 691315 & 0.791 & 4  & 0.666 & 4  & 0.117 & 12 & 0.164 & 12 & 8.00 \\
9  & McMaster NLP     & 694189 & 0.770 & 6  & 0.761 & 6  & 0.208 & 10 & 0.255 & 10 & 8.00 \\
11 & CSE\_IIT\_Ropar  & 694311 & 0.688 & 9  & 0.812 & 9  & 0.242 & 8  & 0.306 & 8  & 8.50 \\
12 & ULL              & 694669 & 0.585 & 13 & 0.846 & 11 & 0.262 & 7  & 0.320 & 6  & 9.25 \\
13 & CtbuY            & 691446 & 0.615 & 12 & 0.848 & 12 & 0.232 & 9  & 0.317 & 7  & 10.00 \\
\bottomrule
\end{tabular}
}
\end{table*}


\begin{table*}[t!]
\centering
\caption{Summary of DreamerNLplus Submissions on Task 3.1 Across Evaluation Metrics (↑ higher is better, ↓ lower is better).
Detail score refers to 
Fig. \ref{fig:task-3.1test-all-teams}
}
\label{tab:dreamer_submissions}
\begin{tabular}{lccccc}
\hline
\textbf{Submission} & \textbf{CS ↑} & \textbf{CT ↓}
& \textbf{ROUGE} & \textbf{BERT} & \textbf{Score Avg} \\
\hline
693964 & medium-high & high & decent & decent & \textbf{0.533 (best)} \\
694011 & medium-high & medium & lower & medium & 0.508 \\
694142 & \textbf{very high/3rd} & \textbf{very low/best} & low & low & 0.404 (worst) \\
\hline
\end{tabular}
\end{table*}

\begin{figure*}[t!]
\centering 
\includegraphics[width=.9\textwidth]{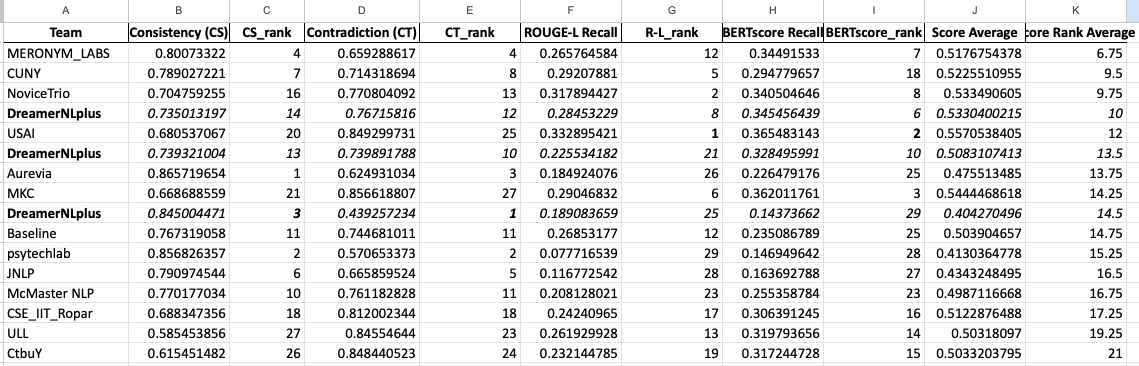}
  \caption{Task 3.1 eval on test set - all teams (our filtering: by using the best averaged rank submission from other teams).
  } 
  \label{fig:task-3.1test-all-teams}
\end{figure*}

\begin{table*}[t]
\centering
\caption{Task 3.2 official rankings for recurrent dynamic signatures of Improvement and Deterioration. Best scores in each metric column are shown in bold, and second-best scores are underlined.
We \textbf{won Improvement} category, and \textit{2nd best} on Specificity for Deterioration category.}
\label{tab:task32_results}
\begin{tabular}{llccccc}
\hline
\textbf{Direction} & \textbf{Team} & \textbf{Rank} & \textbf{Fit} & \textbf{Recurrence} & \textbf{Specificity} & \textbf{Overall} \\
\hline
\textbf{Improvement} & \textit{DreamerNLplus} & 1 & 0.6250 & \underline{0.8125} & \textbf{1.0000} & \textbf{0.7608} \\
 & CSE\_IIT\_Ropar & 2 & \textbf{1.0000} & 0.6875 & 0.3750 & \underline{0.7426} \\
 & MKC & 3 & \underline{0.7500} & 0.5625 & \underline{0.9375} & 0.7266 \\
 & McMasterNLP & 4 & 0.6875 & 0.3750 & 0.7500 & 0.5938 \\
 & psytechlab & 5 & 0.6875 & \textbf{1.0000} & 0.2500 & 0.5437 \\
 & Aurevia & 6 & 0.3750 & 0.6250 & 0.5000 & 0.4653 \\
 & MeronymLabs & 7 & 0.2500 & 0.2500 & 0.5625 & 0.2981 \\
 & CtbuY & 8 & 0.2500 & 0.2500 & 0.0000 & 0.1250 \\
 & CUNY & 9 & 0.0000 & 0.0000 & 0.2500 & 0.0000 \\
\hline
\textbf{Deterioration} & CSE\_IIT\_Ropar & 1 & \underline{0.8750} & 0.5625 & \textbf{0.9375} & \textbf{0.7891} \\
 & Aurevia & 2 & 0.6875 & \underline{0.8125} & 0.5625 & \underline{0.6761} \\
 & \textit{DreamerNLplus} & 3 & 0.4375 & 0.6875 & \underline{0.8750} & 0.6038 \\
 & MeronymLabs & 4 & 0.4375 & 0.5625 & 0.8125 & 0.5511 \\
 & CtbuY & 5 & \textbf{1.0000} & \textbf{1.0000} & 0.0000 & 0.5000 \\
 & psytechlab & 6 & 0.6250 & 0.6250 & 0.2500 & 0.4911 \\
 & MKC & 7 & 0.1875 & 0.1875 & 0.5000 & 0.2301 \\
 & McMasterNLP & 8 & 0.1875 & 0.1875 & 0.3125 & 0.2109 \\
 & CUNY & 9 & 0.0000 & 0.0000 & 0.3125 & 0.0000 \\
\hline
\end{tabular}
\end{table*}

\end{document}